\documentclass[11pt,letterpaper,twocolumn]{article}

\newcommand{\workingDate}{\textsc{April 2, 2014}}
\newcommand{\userName}{Dartmouth College}
\newcommand{\institution}{Dartmouth College}
\usepackage{research}

\begin{document}

\logo

\title{Classics Report}

{\Huge \textcolor{darkblue}{Automated\\[2mm]Attribution and\\[2mm] Intertextual Analysis}}\\[1mm]

\small

\noindent
\textbf{Authors:} James Brofos, Ajay Kannan, and Rui Shu

\noindent
\textbf{Advisor:} Pramit Chaudhuri

\noindent
\textbf{Abstract:}
\textit{In this work, we employ quantitative methods from the realm of statistics and machine learning to develop novel methodologies for author attribution and textual analysis. In particular, we develop techniques and software suitable for applications to Classical study, and we illustrate the efficacy of our approach in several interesting open questions in the field. We apply our numerical analysis techniques to questions of authorship attribution in the case of the Greek tragedian Euripides, to instances of intertextuality and influence in the poetry of the Roman statesman Seneca the Younger, and to cases of ``interpolated'' text with respect to the histories of Livy.}

\setlength\parskip{0.0in}

\tableofcontents

\setlength\parindent{1em}

\section{A Classical Introduction}

Author attribution and intertextual analysis are highly active areas of focus within the study of Classics.  Identifying the author of a work is a fundamental and important facet for analysis.  Determining instances of intertextuality can be seen as a generalization of the author attribution problem, where the primary focus is to determine the existence and extent of influence between works by different authors.  Influence can take many forms, ranging from content to nuances in style.  By establishing the origins of a piece of text, we can better understand the impetus and meaning behind work belonging to the Classical era.

We consider three different ways in which texts may diverge from their claimed, original sources.  In some cases, an entire work may be written by an imposter.  In other cases of textual variants, an author's original work may be adapted by an imposter, characterized by changes in style or wording on a macroscopic scale.  Finally, we consider ``interpolated'' texts, in which text written by other authors is inserted into specific parts of the original work.  Intertextuality also takes many different forms, from similarity in diction to mimicry of style.  Because an author's style is complex and nuanced, we do not seek to categorize different forms of intertext in this paper.  Instead, we attempt to identify a subset of cases of intertextuality and propose scores of similarity based on specific representations of the texts.

This work represents a collaboration between the study of the Classics and the study of computer science. As a result, it is crucial to give perspective to both sides. To the computer scientist, we now take the opportunity to provide insights into the lives and times of those figures from classical antiquity which represent the subjects of our case studies. 

In each of our case studies -- the tragedies of Euripides, the stoicism of Seneca, Livy's histories -- there presents an opportunity to apply quantitative techniques to provide at least one variety of answer to questions of authorship and influence. For each instance in antiquity, we present a brief historical context as well an explanation of the problem that we will seek to address through quantitative techniques.


\subsection{The Tragedy of Euripides}

Euripides was born in the year 480 BC to parents Mnesarchus and Clito on the Greek island of Salamis. On the very day that Euripides was born, Xerxes, the Persian king, was locked in fierce battle with his Greek rivals on the island. It is said that the first successful defense of the island against the Persians came at the site of Euripus, after which Euripides was named.  

Although Euripides' father urged his son to pursue athletics at an early age, there is evidence to suggest that the young Euripides chose instead to undertake the crafting of a dramatic opus. As a pupil of the acclaimed orator Prodicus, Euripides began to develop skills in rhetoric. A poet specializing in works on tragedy, Euripides competed in the Dionysia, the prestigious Athenian Dramatic Festival. However, he was mocked and ridiculed by comic poets of the time, as tragic poets often were by comic poets of the time. Nevertheless, he was widely regarded as a major tragedian and received accolades for several of his plays during his life. After his death, interest in Euripides' works grew rapidly.

With Euripides' plays rising in value after his death, great haste was made to preserve his works along with those of Aeschylus and Sophocles. But with time, significant portions of his work have either been lost or corrupted. While copyist and scholars have since made considerable strides to recover and preserve some of his plays, efforts to construct original copies of his plays continue to be in vain. Corrupted and inconsistent, some of Euripides' plays have garnered much controversy; for some, the original work lies buried beneath endless edits made over the years, for others, the very authorship of some of his plays are in heavy dispute.

\begin{figure}[ht]
\centering
\includegraphics[height=5cm]{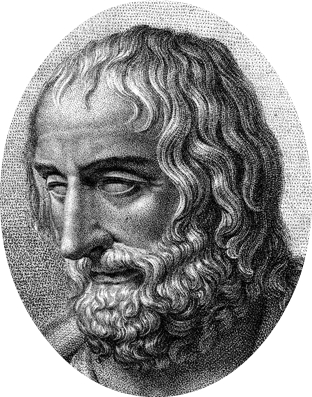}
\caption[8pt]{\small Some of Euripides' plays were victims of heavy modifications. The authorship of certain works, such as \emph{Rhesus}, are still debated. Refer to \cite{euripides_image}.}
\label{fig:euripides}

\end{figure}

\subsubsection{Problem Statement: Euripides}

Euripides' play \emph{Rhesus} has been the subject of scholarly debate since the $17^{\text{th}}$ century due to its questionable authorship claims. In order to capture the stylistic qualities of language, we turn our attention to the representations induced by functional $n$-grams as well as representations induced by the \emph{term frequency-inverse document frequency}. These representations capture ideas of passage ``sound'' as well as word choice so as to allow a learning algorithm to distinguish between Euripides' own language and the language of an imposter.

An additional problem under consideration in the realm of Euripides is that of the play \emph{Iphigenia at Aulis}. This particular work was produced after Euripides' demise and as a result was modified substantially by the author's disciples. This poses an interesting question: ``Will a learning algorithm be capable of identifying a heavily modified work by Euripides from its purer form?''

\subsection{Seneca the Hypocrite}
The vastly popular philosopher and dramatist Seneca was born as Lucius Annaeus Seneca to Helvia and Lucius Annaeus Seneca ``the Elder.'' Seneca's training in rhetoric and his wealthy upbringing brought him
in close proximity to the Imperial court. However, Seneca's connection to the imperial court was wrought with troubles. He was nearly sentenced to death by emperor Caligula and then banished to Corsica
shortly after Claudius came to power, only to be recalled by Claudius's fourth wife Agrippina the Younger to serve as tutor to her son Nero. With Nero's rise to emperor, Seneca would become Nero's advisor along side Sextus Afranius Burrus. But over time as Nero grew, his
influence as advisor waned. Following Burrus' death and amidst accusations of embezzlement, Seneca retired to pursue his studies and writings.

His writings reflected his Stoic philosophy. Despite the popularity of his work in subsequent generations, Seneca himself was heavily criticized by his contemporaries for his failure to abide by his own preachings: Seneca denounced tyranny, and yet served as
advisor to the tyrant Nero; he decried power, but was closely connected to the imperial court; he condemned flattery, and yet wrote many pleas to Claudius for restoration during his exile; he criticized the rich, and yet amassed much wealth himself and lived in luxury.

Nevertheless, his writing was popular after his death. Translations of his works became commonplace, and his tragedies were widely read in medieval and Renaissance European universities.

\subsubsection{Problem Statement: Seneca}
Of Seneca's numerous plays, two particular tragedies are often called into question, namely \emph{Hercules Oetaeus} and \emph{Octavia}. Scholars have now reached the consensus that \emph{Octavia}, despite closely resembling Seneca's style, was written shortly after Seneca's death by someone with a keen knowledge of Seneca's plays and philosophical work. The authorship of \emph{Hercules Oetaeus}, however, is still under dispute.

\subsection{Livy, The Editor of History}

Among historians who have left indelible marks in the annals of history, few have been as prolific and influential as the Roman historian Titus Livius (Livy). Born shortly before the Roman civil wars, Livy lay witness to the fall of Marcus Antonius and the rise of Octavian. As a resident of the city of Patavium caught in the middle of a great civil strife, Livy was undoubtedly poised to make his name through politics or warfare.

Livy's greatest work, \emph{Ab Urbe Condita Libri}, was magnificent in its breadth and influence; Livy's historical work not only spanned the period from the foundations of Rome in 753 BC to the reign of Octavian in Livy's own time, but it was also well-received by his contemporaries and left long-lasting impacts on the style and philosophy of historical writing. Very little is actually known about Livy's life, but given the significance of his work, it is this very mystery that incites scholastic curiosity. 

\begin{figure}[ht]
\centering
\includegraphics[height=6cm]{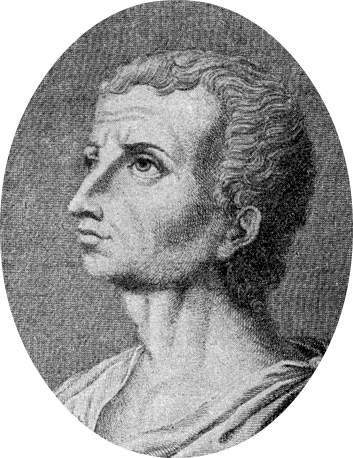}
\caption[8pt]{\small Despite the popular nature of Livy's work, the validity of his quotations are often left unchallenged. Since Livy lacked the political standing necessary to access much of Rome's historical records when he wrote \emph{Ab Urbe Condita Libri}, it was inconceivable how Livy could have amassed the body of records that his historical work would have required. Refer to \cite{livy_image}.}
\label{fig:hierarchical}
\end{figure}

\subsubsection{Problem Statement: Livy}

Livy's work suggested that he was a historical populist.  His charm, like so many of the authors whose work has survived thus far, lies in its ability to evoke the spirit of the times and the soul of the biographer.  His is not the place of a dry encyclopedist - but rather that of a spinner of legend based on fact, but using art to bring the personalities he writes about to life. However, the question remains: ``To what extent did Livy modify the dialogue?'' And, if he did, can we, in fact, see hints of Livy's voice underlying the quotes he used. 

To answer this question, we turn our attention again to the problem of textual representation. For this particular application, we believe that the function $n$-gram technique to be of particular interest because of its ability to represent the way that a passage  fundamentally \emph{sounds}. Ultimately, what we are after is measure of how ``Livian'' a passage sounds.

\subsection{Intertextuality}

\section{Computational Methods and Formulations}
\label{sec:methods}

\subsection{Textual Representation Methods}

\subsubsection{Functional \texorpdfstring{$n$}{n}-Gram Representations}

Developed by Forstall and Jacobson \cite{fng}, the functional $n$-gram is a sequence of $n$ characters that captures phonetic qualities of texts.  Forstall and Jacobson applied this sound-oriented approach with encouraging results to analyze intertext between Paul the Deacon's \emph{Angustae Vitae} and works by Catallus.  Because of the importance of sound in the Classics, we chose to employ functional $n$-gram probability features in our work as well.

The probability features are built according to the following function:

\begin{equation}
P(e_n | e_{n-N+1}^{n-1}) = \frac{C(e_{n-N+1}^{n-1}e_n)}{C(e_{n-N+1}^{n-1})}
\end{equation}

\noindent 
where $e_{n-N+1}^{n-1}$ is the substring of text from characters $n-N+1$ to $n-1$, $e_n$ is the $n^{th}$ character in text, and $N$ is the length of the letter gram.

\subsubsection{Term Frequency-Inverse Document Frequency}

As an alternative to the functional $n$-gram representation, we also implement term frequency-inverse document frequency (\emph{tf-idf}) as an additional representation over documents. 

The purpose of \emph{tf-idf} is to identify ``important'' words in a document.  While frequency is an obvious factor in term importance, common words such as prepositions, pronouns, and conjunctions would receive a high score when in fact they are not central to the document.  Thus, $tf-idf$ lowers the score of words that appear often in a set of texts, the corpus.

Formally, \emph{tf-idf} is the product of two functions, term frequency (\emph{tf}) and inverse document frequency (\emph{df}), as shown below:

\begin{align}
	\mathcal{F}\left(t,d,C\right) = f(t,d) \times d(t,C)
\end{align}

where $d(i,C) = log\frac{|D|}{1 + |\{d \in C: t \in d\}|}$, $f$ is the frequency function, $t$ is the term, $d$ is the document, and $C$ is the corpus to which $d$ belongs. Taken together, $\mathcal{F}\left(\cdot,\cdot,\cdot\right)$ gives the \emph{tf-idf} representation.

\subsubsection{Probability Distributions over Representations}

By adopting a probabilistic perspective of the features, we achieve a more generalizable representation space. The only assumption that is necessary to make is that the distribution of examples in $\mathcal{D}_{\text{passage}}$ are an appropriate approximation to the true distribution of $\mathcal{D}$. By using a probabilistic representation of the features, we find that we are able to more accurately identify differences in style.  This is likely because probability features are more robust to differences in passage length than raw counts.

\subsection{Pseudo-Metrics}

\subsubsection{Bhattacharyya Distance}

When attempting to quantify the similarity of textual features between two authors, it is often useful to compare the distribution of each author's features over the entire feature space.  The Bhattacharyya distance is a measure that does just this.  The definition of Bhattacharyya distance $D_B$ for discrete probability distributions $p,q$ over $X$ is as follows:

\begin{align}
D_{\text{Bhattacharyya}}\left(p,q\right) = -\log \left[BC(p,q)\right]\\
\text{where } BC(p,q) = \sum_{x \in X} \sqrt{p(x)q(x)}.
\end{align}

The Bhattacharyya distance was used in this study over the Mahalanobis distance pseudo-metric because the Mahalanobis distance disregards differences in standard deviation of distributions with similar averages, which is likely the case as a result of common Latin prepositions, conjunctions, and articles being used often across texts.

\subsection{Statistical Inference and Prediction Algorithms}

\subsubsection{One-Class Support Vector Machine}
We use of the one-class support vector machine (SVM), a learning algorithm for anomaly detection, in author attribution and intertextuality analysis.  By training the one-class SVM using undisputed texts by one author, we learn a frontier between that author's writing style and that of other authors.  Thus, this algorithm can be used to provide a classification decision as to whether a piece of text is written by a specific author or not.

Developed by Sch{\"o}lkopf in 1999, this algorithm solves the following quadratic program to find the maximum distance of the training set data points from the origin:

\begin{align}
\operatorname*{min}_{w \in F, \xi \in \mathbb{Z}^l, \rho \in \mathbb{R}} \frac{1}{2} \lVert{w}\rVert^2  + \frac{1}{vl} \sum_{i} \xi_i - \rho\\
\text{subject to: } \langle w \cdot \Phi(x_i) \rangle \geq  \rho - \xi_i\\
\text{ where }  \xi_i \geq 0
\end{align}

\noindent where $w$ and $\rho$ are the weight vector and offset for the hyper-plane in feature space $F$, $\xi_i$ are ``slack'' variables that penalize the objective function for outliers, $\Phi$ is the kernel function, $l$ is the length of the input vector $x$, and $v \in (0, 1]$ upper bounds the proportion of the input data classified as outliers.

The prediction function then takes the form:

\begin{equation}
f(x) = \operatorname{sgn}(\langle w, \Phi(x) \rangle - \rho).
\end{equation}

In practice, the dual form, shown below, using introducing Lagrange multipliers is used to solve the quadratic program.

\begin{align}
\operatorname*{min}_{\alpha} ~~\frac{1}{2} \sum_{i,j}\alpha_i\alpha_jk(x_i, x_j)\\
\text{subject to: } 0 \leq \alpha_i \leq \frac{1}{vl}\\ \sum_{i=1}^{l}\alpha_i = 1,
\end{align}

\noindent where $i,j \in \{1, 2, 3, ..., l\}$ and $\alpha$ contains the Lagrange multipliers.

We elected to utilize the Gaussian Radial Base Function (RBF) as a kernel after witnessing that a polynomial kernel of degree three severely overfit the data, most likely exacerbated by the small dataset. Moreover, the RBF kernel projects the data into a hypothetically infinite-dimensional polynomial space, and so it is more expressive of separability properties of the data than is a finite-dimensional polynomial kernel.  The parameter $v$ was experimentally derived using grid search, as were as kernel function parameters.

\subsection{Bayesian Statistics and A Novel Method for Estimating Occurrences of Intertextuality}

Determining how often intertextual passages are found between texts written by different authors is an indicator of the level of influence one author had on the other.  In algorithms and metrics used to identify intertextuality between documents, parts of one document must inevitably be compared to parts of the other.  However, a ubiquitous problem in string comparisons made with uncertainty is that it is computationally expensive to match the $n-m$ possible substrings in a document against a passage of length $m$, where $n$ is the character length of the document. For document to document comparisons, we expect to have to perform large numbers of time consuming string comparisons.  For these reasons, we present a method to estimate the number of intertextual passages in a document given a query passage and an error tolerance with fewer substring comparisons.

First, we wish to define the following term:
\begin{description}
\item[Admissible Intertextuality:] Suppose at the $i^{\text{th}}$ character in a document we have the beginning of a substring of length $m$ that demonstrates intertextuality according to our metrics. Then almost certainly the substring beginning at the $i+1^{\text{st}}$ character will also demonstrate intertextuality. However, this instance is inadmissible because the intertextuality being recognized is no longer novel, and we have wasted computation detecting the same instance. Admissible intertextuality is defined to be the closest instance of intertextuality from a position $i$ that is not in reference to the same intertextual object.
\end{description}

We also define the parameter $s$ which is the ``skipping distance'' from position $i$ to the next closest instance of admissible intertextuality. That is, is we have evaluated the substring of length $m$ at position $i$ and the next instance of admissible intextuality occurs at the $i+j^{\text{th}}$ position, then $s=j$. Our purpose that is to develop an estimator $\hat{s}$ of $s$.

Suppose that any $m$-character substring of an $n$-character document has constant prior probability $p$ of being intertextual with a query passage.  Consider randomly sampling $b$ substrings of the document for intertext.  As $n$ increases, the true proportion of intertextual substrings in the document will approach $p$; that is, $\frac{\mu}{n}\to p$ as $n\to\infty$, where $\mu$ is the number of substrings of length $m$ that truly contain intertextuality. Hence, whether the substrings contain intertext can be modeled as a series of $b$ Bernoulli trials $X_1,X_2, X_3,\ldots, X_b$ with probability $p$.  Thus, by Hoeffding's inequality,
\begin{align}
\mathbb{P}\left[\vert\overline{X}_b - p\vert > \epsilon\right] \leq 2e^{-2b\epsilon^2}
\label{eq:hoeffding}
\end{align}
for any $\epsilon > 0$. Here $\overline{X}_b = b^{-1}\sum_{i=1}^b X_i$.

Under our construction, this inequality implies that as the number of substrings chosen increases linearly, the likelihood we miscalculate the number of intertexual references given a document and query passage decreases exponentially.  By setting a tolerance for the maximum acceptable deviation of the detected proportion of intertextuality to the true proportion, we can determine the number of sample substrings necessary so as to achieve a desired probabilistic bound on the right-hand side.

In practice, we allow $\epsilon \approx 0.1$. This implies that the amount of deviation from the proportion of observed intertextual instances to the true proportion is unlikely to exceed $0.1$ as $b$ becomes larger. Suppose we impose an upper bound of $\alpha$; that is, $\mathbb{P}\left[\vert \overline{X}_b - p\vert > \epsilon\right]\leq \alpha$. Then
\begin{align}
\epsilon = \sqrt{\frac{1}{2b}\log \left(\frac{2}{\alpha}\right)} = 0.1\\
\implies b = 50\log\left(\frac{2}{\alpha}\right)
\label{eq:n_samples}
\end{align}
This construction allows one to induce a kind of confidence interval of ninety-five percent on the estimated proportion $\overline{X}_b$ as $\mathcal{C} = \left(\overline{X}_b-\epsilon, \overline{X}_b + \epsilon\right)$ and $\mathbb{P}\left[p\notin \mathcal{C}\right] = \mathbb{P}\left[\vert\overline{X}_b-p\vert > \epsilon\right] \leq \alpha$.

We opt to evenly space out the samples throughout the passage to maximally achieve the independence criterion of the construction above.  Thus, a sample will occur every $\floor*{\frac{n-m + 1}{s}}$ characters.  When the length of the intertextual passage is less than $\floor*{\frac{n-m + 1}{s}} - m$, where $n$ is the document length, $s$ is the number of samples, and $m$ is the query length, then we are guaranteed that an intertextual passage will be present in no more than one sample.  From equation (12), we see that the number of samples does not depend on the passage length.  Hence, when comparing query passages of reasonable length to entire documents, the independence criterion between samples will be met in the vast majority of cases.

\section{Oracle of DelPy: An Intertextual Tool}

In order to promote further research among classicists in the application of quantitative techniques to textual analysis, we have produced \delpy. The \delpy~software represents a Python implementation of the statistical inference and prediction algorithms and representations detailed in \cref{sec:methods}. 

In the development of \delpy~we sought to emphasize two programming principles. First, we believed that the software should be accessible to non-technical researchers, and therefore our software implements pre-packaged analysis techniques that should \emph{just work} given textual input. Second, we adhere to the paradigm of object-oriented programming. As a result, we implement many classes for representing documents in a corpus and for performing textual representation and analysis on those representations.

Let us seek to pin down in this section the structure of \delpy~and why the software represents an excellent resource for classicists seeking to engage with quantitative textual analysis. The software may be freely obtained from \url{https://github.com/JamesBrofos/DelPy}.

\subsection{Software Structure}

\begin{description}
\item[Document Representation:] To begin computational analysis of pieces of literature, the raw texts downloaded from online sources must be preprocessed and made into lists of words.  Certain characters, including line numbers and non-ascii characters, are removed, and the text is stored both as a list of words and one large processed string.  The user can choose how to handle punctuation and word capitalization, but for ease of use, defaults are included in the software.  All preprocessing functions are handled upon creation of a Text object to represent a text.  All functions have linear time complexity in the length of the raw text in characters, providing robust scalability.

Because a significant number of Classical texts do not employ a Roman alphabet (Euripides and subsequent Greek authors, in particular), we also develop a \texttt{GreekText} document representation class that transforms ascii-standard Ancient Greek to a format interpretatable by our textual representation objects. (Subsequently, it was necessary to implement a \texttt{GreekFunctionalNGram} class.)

\item[Textual Feature Representation:] The method for representing a document is written mathematically as $f : \mathcal{D}_{\text{document}} \to \mathcal{X}$ such that $f(D) = X$, where $D$ is a document and $X$ is the corresponding features in space $\mathcal{X}$ of that document. 

We believe that fully representing a document should be as programmatically simple as $f(D) = X$. Therefore, after creating a \texttt{Text} object, the user has the choice of representing the text as functional n-grams or using term frequency-inverse document frequency (\emph{tf-idf}). Simply by creating a \texttt{FunctionalNGram} object with the \texttt{Text} object as the only parameter, functional bigrams and their respective probability features are computed.  The user has the option to specify different length functional $n$-grams when constructing the object if he or she wishes.  Functional $n$-gram frequencies and probability features are stored as dictionaries, with the $n$-gram as the key, enabling the user to easily query the text for specific information.  Again, all functions are linear in the length of the text in characters.

To obtain \emph{tf-idf} information for a text, the user must first create the object and add Text objects to the corpus.  Upon adding each texts, the term frequencies are updated.  The user can call functions to obtain similarity scores and obtain the features for the corpus as input for prediction functions elsewhere in the software package.  All functions are $\mathcal{O}\left(n\right)$, where $n$ represents the number of words in the corpus.

\item[Intertextuality Module:] A major component of this work lies in the automatic identification of passages in two separate documents that may exhibit a degree of similarity in style, wherein one author has visibly influences the other. This phenomenon is referred to as ``intertextuality.'' In \delpy~we make several simplifying assumptions so as to make the intertextuality problem tractable for automation:
\begin{enumerate}
\item We assume that the length of passages in which intertextuality exists are approximately equal. In particular, if $s_n$ is a passage of length $n$, then the set of all passages in an accompanying work that display intertextuality is denoted $\left\{s_n^{(i)}\right\}_{i=1}^m$; that is, all $m$ passages have equal length $n$. 
\item We assume that intertextuality may be detected via a metric or a pseudo-metric give a representation of the passage. That is, if $\mathcal{X}$ is a feature space capable of capturing intertextuality, then a metric (or pseudo-metric) $\mathcal{M}$ has the property that for two intertextual passage representations $x_n^{(1)}$ and $x_n^{(2)} \in\mathcal{X}$, $\mathcal{M}\left(x_n^{(1)},x_n^{(2)}\right)\to 0$ as the representation becomes more expressive. In practice, however, it is necessary to establish a thresholding parameter rather than find passages with a distance equal to zero.
\item We assume that the user of the software is capable of producing a passage that she believes may be reproduced in an intertextual sense in another document. Given this passage and the corresponding document, the software seeks to find subsequent passages that display intertextuality according to the previous two assumptions. 
\end{enumerate}

We should stress that it is presently intractable to allow the user to specify either variable length instances of intertextuality or two documents to fully parse for instances of intertextuality. Permitting either of these conditions would grow the computational complexity of the intertextual calculations to a many-to-many relationship. By comparison, the assumptions we specified previously restrict the calculations to a one-to-many relationship. 

The search for intertextuality is implemented in the \texttt{IntertextualityModel} module in \delpy. The \texttt{IntextualityPassage} class allows the user to specify \texttt{Text} objects representing both the passage and the document to query. Upon instantiation, the \texttt{IntextualityPassage} object calculates an intertextuality metric for every substring of length $n$ occurring in the query document. This process operates in $\mathcal{O}\left(n\right)$ time, but in practice it is quite computationally expensive and searching a lengthy corpus for instances of intertext  will take time because the number of comparisons performed is $\mathcal{O}\left(nm\right)$, where $n$ is the length in characters of the document and $m$ is the passage length.

\subsection{Document-to-Document Intertextuality Comparisons}

We leverage our implementation of passage-to-document comparison in document-to-document comparisons by iterating over the first document for passages of various sizes and querying for similar passages in the second document. However, due to the time complexity involved with this approach, we decided to query only parts of the original text within the second text.

Considering that the complete analysis is not feasible within short time frames, the simplest approach for a user who is looking to formulate novel hypotheses on connections between two documents is to obtain arbitrary samples of intertextuality between the two documents.  Thus, we randomly sample $n\times l$ passages from the first document, where $n$ is the number of passages to sample at each substring length and $l$ is the number of substring lengths to check between a maximum and minimum size for query passages. The parameter $n$ and the variables that define $l$, namely maximum and minimum intertextuality query sizes and step-size, are all inputs into the \texttt{IntertextualityDocumentComparison} class, with defaults provided for ease of use.

For each sample, we create an \texttt{IntertextualityPassage} object to compare the query passage to samples in the second document, with the number of samples defined by \cref{eq:hoeffding,eq:n_samples}.  We consolidate the results and subsequently return the results to the user.

\end{description}

\section{Automatic Textual Analysis with Applications to Classics}

\subsection{Case Study: Euripides}
We first apply our one-class SVM and functional $n$-gram probability features approach to determine how similar undisputed Euripidean texts are to \emph{Iphigenia at Aulis} and \emph{Rhesus}.  While \emph{Rhesus} is alleged to have have been written by an imposter, \emph{Iphigenia at Aulis} is hypothesized to be a heavily-modified Euripidean work.  We wish to not only obtain a binary classification of these two texts, but also to obtain measures of similarity between \emph{Iphigenia at Aulis} and true Euripidean texts to estimate the extent to which the disputed text may have been modified.

Non-disputed Euripidean tragedies, romantic dramas, and political dramas were used to train the one-class SVM, namely \emph{Electra, Helen, Heracles, Hercuba, Ion, Iphigenia at Tauris, Orestes, Phoenissae, Suppliants, Trojan Women}, and \emph{Bacchae}.  The one-class SVM classified both \emph{Iphigenia at Aulis} and \emph{Rhesus} as non-Euripidean texts.\\

\begin{figure}[ht]
  \centering
  \includegraphics[width=0.5\textwidth]{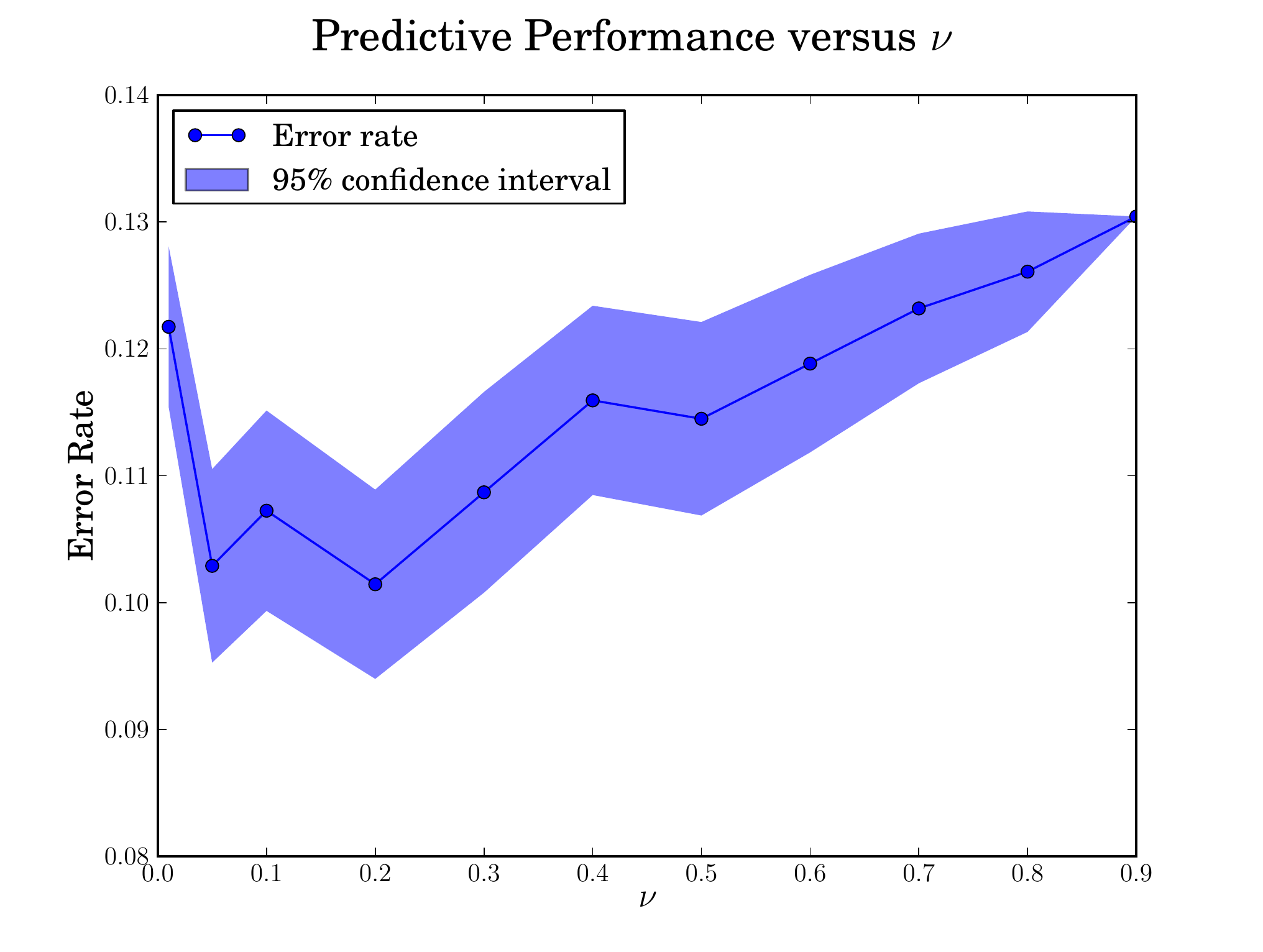}
  \caption{The parameter $\nu$ controls how strictly the decision boundary fits the training dataset.  We experimented with different values of $\nu$ to examine optimal settings for the one-class SVM.}
\end{figure}

\begin{center}
\captionsetup{type=table}
\begin{tabular}{ | p{3cm} | p{4.5cm} |}
  \hline
  \textbf{Text} & \textbf{Signed Distance from Hyper-plane} \\ \hline
  \emph{Iphigenia at Aulis} & -0.334 \\ \hline
  \emph{Rhesus} & -0.488 \\ 
  \hline
\end{tabular}
\captionof{table}{Note that the negative values denote that the texts were both classified as stylistically different from true Euripides texts.}
\end{center}

While the one-class SVM does not return posterior probabilities involved in classification, a metric that can be extracted is the distance of a data point to the hyper-plane.  In general, points that lie farther from the hyper-plane are more anomalous than those that are not.  An encouraging result from this experiment is that \emph{Iphigenia at Aulis} is closer to the hyper-plane than \emph{Rhesus}.  Meanwhile, points in the training set are on average within $10^{-2}$ units of the hyper-plane.  In sum, our model lends evidence towards the two disputed texts being written or heavily influenced by other authors and seems to register Euripides' influence on \emph{Iphigenia at Aulis}.

We also attempted using \emph{tf-idf} features as input to the one-class SVM, both alone and in conjunction with functional $n$-grams.  Using \emph{tf-idf} features alone yielded poor results, which suggests that sound-based stylistic qualities are more powerful indicators for author attribution and intertext than word choice alone.  When \emph{tf-idf} features were appended to functional $n$-gram probability features, the one-class SVM performed very poorly.  Because the training set only included ten data points, we believe this outcome was the result of severe overfitting.

\subsection{Case Study: Seneca}

As mentioned in the introduction, there are authorship disputes surrounding \emph{Hercules Oetaeus} and \emph{Octavia}, two out of the ten ``Senecan'' tragedies.  While classicists are not in agreement as to whether \emph{Hercules Oetaeus} was written by Seneca or not, it is widely accepted that \emph{Octavia} was written and published by an imposter who mimicked Seneca's style after Seneca's death.  The results of our one-class SVM with functional $n$-gram probability features may provide additional insight that will move the classics community closer to settling the author attribution questions surrounding \emph{Hercules Oetaeus}.  In addition, we hope that our model correctly identifies \emph{Octavia} as a false Senecan text, which would lend additional credence to our approach.\\

\begin{figure}[ht]
  \centering
    \includegraphics[width=0.5\textwidth]{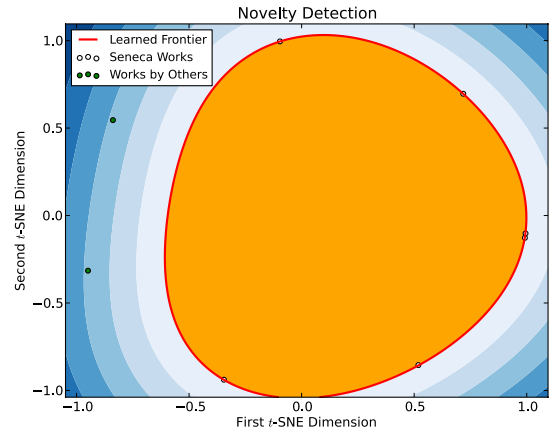}
    \caption{The output of the first experiment, training the one-class SVM on all eight undisputed Senecan tragedies and testing on the two disputed texts.  $t$-SNE was used to project the high-dimensional space into two dimensions.}

\end{figure}

The initial experimental setup was to train the SVM using all eight non-disputed Senecan texts and predict class membership using the other two tragedies.  The model correctly identifies \emph{Octavia} as non-Senecan and also classifies \emph{Hercules Oetaeus} as non-Senecan.  An interesting observation is that \emph{Octavia} is closer to the hyper-plane than \emph{Hercules Oetaeus}.  While \emph{Octavia} was written by an imposter, classicists agree that the imposter was highly skilled and well-versed in Seneca's style.  Our preliminary results seem to support this claim.

\begin{center}
\captionsetup{type=table}
\begin{tabular}{ | p{3cm} | p{4.5cm} |}
  \hline
  \textbf{Text} & \textbf{Signed Distance from Hyper-plane} \\ \hline
  \emph{Hercules Oetaeus} & -0.126 \\ \hline
  \emph{Octavia} & -0.008 \\ 
  \hline
\end{tabular}
\captionof{table}{The negative values denote that the texts were classified non-Senecan.}
\end{center}

To validate our model, we included various non-Senecan texts in the test set, including works by Cicero, Livy, Virgil, Lucan, and Ovid.  Out of 20 non-Senecan texts, all but two were correctly classified by our model:
Lucan's \emph{Pharsalia} and Ovid's \emph{Metamorphoses}.  Lucan was in fact Seneca's nephew, and the heavy influence of Seneca on Lucan's work may explain the misclassification.  In addition, Ovid had heavy influence on Seneca's writing, which may explain the second misclassification.  What may be of interest is that while Ovid's work is similar to Seneca's works, Virgil's works were not classified as Senecan.  This result suggests that Ovid was a more direct influence on Seneca than was Virgil.  Relative influence of Ovid and Virgil on Seneca is still an open question, and these results put forth an interesting hypothesis to explore both via traditional classics approaches and more computational methods.

As a final experiment, we attempted to use fewer Senecan tragedies in the training set.  Using only four Senecan texts led to unacceptably high error rates, but using six Senecan tragedies had slightly better results.  There were $\binom{8}{6} = 28$ combinations, keeping the two disputed texts out of the training set.  The following error matrix depicts the tendency of our model to categorize true Senecan texts as non-Senecan.\\
 
\begin{center}
\captionsetup{type=table}
\begin{tabular}{ | c | l | l | l | l |}
  \hline
  & \multicolumn{4}{|c|}{\textbf{Number of Misclassified}}\\
  & \multicolumn{4}{|c|}{\textbf{False Senecans}}\\ \hline
  & & Zero & One & Two \\ \cline{1-5}
  \textbf{Number of} & Zero & 5 & 0 & 0\\ \cline{2-5}
  \textbf{Misclassified} & One &  11 & 0 & 0\\ \cline{2-5}
  \textbf{True Senecans} & Two & 12 & 0 & 0\\ \hline
\end{tabular}
\captionof{table}{All the misclassification errors were classifying undisputed Senecan texts as non-Senecan.}
\end{center}

Though there were many misclassifications, the true Senecan texts classified as non-Senecan were all much closer to the hyper-plane than the disputed texts, usually within .05 units of the hyper-plane.  Furthermore, this behavior is desirable both for its consistency and for providing more hypotheses for classicists to explore rather than failing to identify interesting results.  The high misclassification rate was most likely because of the small training set.  This experiment was informative in gaining intuition on the number of training set texts our model may need for accurate classification.  However, it is important to realize that a lower bound on an optimal training set size may differ from author to author based on the variability in the author's style from work to work.

As a final experiment, we attempted to include Senecan prose along with Senecan tragedies in the training set for our model.  The result was that many non-Senecan texts were classified as Senecan, most likely due to the high variability in the training set.  This result suggests that the sounds and style of Senecan tragedies vary greatly from his prose.

\subsection{Case Study: Livy}
Livy's notion of a quote in his \emph{History of Rome} tomes is a loose one.  Not clearly demarcated, the quotes in his works are believed to range from near word-for-word transcription to heavily-adapted paraphrasing.  Moreover, classicists suspect that some of the quotes Livy attributes to other authors are in fact written by him.  Until now, the problem of identifying dubious quotes has been intractable because of the sheer number of quotes within Livy's works and the effort involved with analyzing each one thoroughly.  The one-class SVM with functional $n$-gram probability features provides a coarse analysis tool for quotes to identify types of quotes that are more likely to be written by Livy himself.  However, our model faces obstacles in classifying specific quotes as characteristically ``Livian'' or not.  

We begin by examining Livy's quotation style with a broad lens, first by grouping all quotations together, and then grouping quotations by whether they were direct, impersonal, variant, or impersonal variant quotes \cite{wife}.  The database of quotes used was compiled by Andrew Zigler, and more information about quote categorization can be found in \cite{wife}. The one-class SVM was trained using functional $n$-gram probability features for each of Livy's books, with quotes removed.  

In the first experiment, the SVM classified all quotes concatenated together as non-Livian.  In the second, direct quotes, impersonal quotes, and impersonal variants of quotes were all classified as non-Livian, with variant quotations classified as Livian.  These preliminary results corroborate the hypothesis that some of Livy's quotes are in fact heavily doctored and rewritten in his own voice.  The fact that impersonal variant quotes ``sound'' more Livian on average than variant quotes is an interesting result to be explored further by classicists.  The fact that our model correctly identifies direct quotes as non-Livian lends credence to the accuracy of our model. \\

\begin{center}
\captionsetup{type=table}
\begin{tabular}{ | p{4cm} | p{3.5cm} |}
  \hline
  \textbf{Type of Quote} & \textbf{Signed Distance from Hyper-plane} \\ \hline
  Direct & -0.093 \\ \hline
  Impersonal & -0.289 \\ \hline
  Impersonal Variant & -0.147 \\ \hline
  Variant & $~$0.059 \\ \hline
\end{tabular}
\captionof{table}{Negative values denote data points classified as non-Livian and positive values denote data points classified as Livian text.}
\end{center}

While broad analysis was a success, we encountered difficulties in analyzing short quotes.  In the previous case studies we presented, entire texts were used to compute the functional $n$-gram probability features.  However, quote length is orders of magnitude smaller than the length of an entire text.  With such a small sample size for estimating the functional $n$-gram probability features, quotes in the test set are nearly always classified as not stylistically similar to Livy.  Training the one-class SVM on quote-sized chunks of true Livian prose suggested that some of Livy's quotes were written in Livy's style and others were not.  However, we believe the error rate is quite high due to the short lengths of the quotes tested.

Any probabilistic feature drawn from short quotes will necessarily be more uncertain and prone to error in attribution analysis.  Factoring in uncertainty in the probability distributions may be a next step for more finely-tuned passage analysis.  From a macroscopic viewpoint, our results from this case study suggest that while many of Livy's quotes indeed preserve the quoted authors' presentation, some quotes are heavily-doctored or possibly fabricated by Livy himself.  With longer quotes, our method is still a viable method to gather preliminary findings on the true origin of a passage in Livy's texts.

\section{Conclusion}

In this paper, we present algorithms and software for performing authorship attribution and intertextual analysis.  We demonstrate the efficacy of our methods by applying our methods to open problems in the field of classics. The results we obtained both corroborate generally accepted theories by Classicists today and also add computational recommendations to several important and controversial questions that arise from historical antiquity.  These results lend confidence to our tools as both a means for developing hypotheses for Classical theorists and for compiling additional evidence to support claims.  

Because the software we developed is generalized to handle any texts using Latin or Greek alphabets and does not rely on domain knowledge, these tools are applicable to a wide variety of applications in author attribution and intertextual analysis.  We hope to expand the software package as we confirm the validity and efficacy of novel feature representations of the texts via additional case studies.

\subsection{Recommendations for Further Research}

There are several avenues in which this research could progress further. In particular, a exploitation of metrical analysis in the form of the rhythm of poetry may yield an interesting feature upon which to train a classifier. Moreover, this research has purposefully avoided the use of domain knowledge to support the construction of the learning algorithms. Domain knowledge has been successfully leveraged in other areas of research, and we feel that its application to classical study will yield superior, if less general, classifiers.


\begin{thebibliography}{99}

\bibitem{livy_image}
``Bibliothek des allgemeinen und praktischen Wissens. Bd. 5'' (1905), Abriss der Weltliteratur, Seite 50.

\bibitem{wife}
``Citation and the Dynamics of Tradition in Livy's AUC.'' Histos 7 (2013): 21-47.

\bibitem{fng}
``Evidence of Intertextuality: Investigating Paul the Deacon's Angustae Vitae,'' Christopher W. Forstall, Sarah Jacobson, Walter J. Scheirer, Literary \& Linguistic Computing (LLC), September 2011.

\bibitem{peck}
Peck, Harry Thurston. \emph{Harper's Dictionary of Classical Literature and Antiquities}. New York: Cooper Square, 1962. Print.

\bibitem{euripides_image}
``Poetry Walks.'' : \emph{Euripides the Athenian by Seferis}. N.p., n.d. Web. 01 May 2014.

\end{thebibliography}
\end{document}